\documentclass[conference, letterpaper]{IEEEtran}
\usepackage[letterpaper, left=0.625in, right=0.625in, bottom=1.02in, top=0.75in]{geometry}
\IEEEoverridecommandlockouts
\usepackage[style=ieee]{biblatex}
\DeclareNameAlias{default}{family-given}
\usepackage{amsmath,amssymb,amsfonts}
\usepackage[ruled]{algorithm2e}
\usepackage{algorithmic}
\usepackage{graphicx}
\usepackage{textcomp}
\usepackage{xcolor}
\usepackage{subcaption}
\usepackage{makecell}
\def\BibTeX{{\rm B\kern-.05em{\sc i\kern-.025em b}\kern-.08em
    T\kern-.1667em\lower.7ex\hbox{E}\kern-.125emX}}
\usepackage{caption}

\DeclareCaptionFont{mysize}{\fontsize{8}{9.6}\selectfont}
\begin{document}

\title{Experimental Evidence that Empowerment May Drive Exploration in Sparse-Reward Environments\\
\thanks{This work was funded by the Center for Innovation and Leadership of Swarthmore College. The work by MB and RK on this publication was made possible through the support of a grant from Templeton World Charity Foundation, Inc. The opinions expressed in this publication are those of the authors and do not necessarily reflect the views of Templeton World Charity Foundation, Inc.}
}

\author{\IEEEauthorblockN{Francesco Massari}
\IEEEauthorblockA{\textit{Swarthmore College}\\
Swarthmore, USA \\
rmassar1@swarthmore.edu}
\and
\IEEEauthorblockN{Martin Biehl}
\IEEEauthorblockA{\textit{Araya Inc.}\\
Tokyo, Japan \\
martin@araya.org}
\and
\IEEEauthorblockN{Lisa Meeden}
\IEEEauthorblockA{Department of Computer Science \\
\textit{Swarthmore College}\\
Swarthmore, USA \\
meeden@cs.swarthmore.edu}
\and
\IEEEauthorblockN{Ryota Kanai}
\IEEEauthorblockA{\textit{Araya Inc.}\\
Tokyo, Japan \\
kanair@araya.org}
}

\maketitle

\begin{abstract}
 Reinforcement Learning (RL) is known to be often unsuccessful in environments with sparse extrinsic rewards. A possible countermeasure is to endow RL agents with an intrinsic reward function, or `intrinsic motivation', which rewards the agent based on certain features of the current sensor state. An intrinsic reward function based on the principle of empowerment assigns rewards proportional to the amount of control the agent has over its own sensors. We implemented a variation on a recently proposed intrinsically motivated agent, which we refer to as the `curious' agent, and an empowerment-inspired agent. The former leverages sensor state encoding with a variational autoencoder, while the latter predicts the next sensor state via a variational information bottleneck. We compared the performance of both agents to that of an advantage actor-critic baseline in four sparse reward grid worlds. Both the empowerment agent and its curious competitor seem to benefit to similar extents from their intrinsic rewards. This provides some experimental support to the conjecture that empowerment can be used to drive exploration.
\end{abstract}


\section{Introduction}
In Reinforcement Learning (RL), relying on feedback from the environment is effective, if rewards are readily dispensed after every action, but many problems offer only sparse guidance of such form. One way to solve this issue is to complement the extrinsic reward signal, which is received from the environment, with an intrinsic reward signal. The agent is endowed with an internal function that takes as input the current sensor state and rewards the agent according to how desirable it is from an intrinsic perspective. This internal reward thus functions as an \emph{intrinsic motivation} \cite{Oudeyer, formalTheory}. 

Since the key to success in sparse reward settings is exploration \cite{RLbook}, the literature on intrinsic motivations features various approaches to drive exploratory behavior \cite{Oudeyer, formalTheory}. One concept that has been proposed as an intrinsic motivation is \emph{empowerment} \cite{powerOrig, powerIntro}, which makes the agent maximize the control it has over its future sensor states.

In \cite{formalTheory}, it was noted that previous empowerment implementations had assumed accurate internal models of the environment as given and it was suggested that empowerment does not motivate the development of such models through exploration, because empowerment maximization does not imply the maximization of \emph{information gain}. Information gain, (also known as \emph{Bayesian surprise} \cite{Bayes}, which is maximized by \emph{knowledge seeking} agents \cite{nerd}) directly measures improvements of the agent's internal model of the environment. However, it was later speculated in \cite{powerIntro} that, in order to control its future sensor states, the agent has to accurately model the environment and that, therefore, empowerment maximization could also drive the improvement of such models, thereby leading to exploration. This would mean that empowerment maximization should be similarly useful as an exploration drive as information gain. 

We here tested an agent that maximizes an empowerment-inspired intrinsic reward and one that maximizes an information-gain-inspired intrinsic reward. We find some promising evidence that the empowerment-inspired intrinsic reward also motivates exploration. However, further research is needed to establish whether, when it comes to motivating exploration, empowerment can be seen as an alternative to information gain.

\section{Background}
\subsection{Partially Observable Markov Decision Process}
For the present work, learning is conceptualized as a \emph{Partially Observable Markov Decision Process}, which is structured into a sequence of
discrete time steps $t = 0, 1, 2, 3, \ldots$, later referred to as `frames'. At every such time step,
the agent senses a part $s_{t} \in \mathcal{S}$ of the environment state $e_{t} \in \mathcal{E}$. It then performs an action $a_{t} \in \mathcal{A}$, thus
in turn affecting the next environment state $e_{t+1}$. The environment responds
by supplying the agent with a new sensor state $s_{t+1}$,
as well as an extrinsic reward $r_{t+1}^{ex} \in \mathcal{R} \subset \mathbb{R}$ \cite{RLbook}.

\subsection{Intrinsic Motivation} \label{IntrinsicMotivation}
An intrinsic reward function takes in the current sensor state $s_{t}$ and features of the agent's internal architecture and returns a real value $r^{in}_{t}$, which together with the extrinsic reward $r^{ex}_{t}$ makes up the total reward $r_{t} = r^{in}_{t} + r^{ex}_{t}$ \cite{formalTheory}. It is possible to design this intrinsic reward function such that it encourages exploratory behavior \cite{Oudeyer, curious}.

\subsection{Empowerment} \label{power}
Empowerment seeks to quantify the level of control an agent has over its environment. The central idea is that of viewing the environment as an information channel between actions and future sensor states. Via its actions, an agent can influence its environment, which in turn determines the agent's next sensor state. It follows that having control over one's future sensor states requires having some control over the environment. Note that the influence an agent has over its environment only contributes to its empowerment in so far as it is observable by the agent. Empowerment then quantifies this control as the channel capacity (in state $e_{t}$) from actions $A_{t}$ to future sensor states $S_{t+1}$, i.e. the maximum amount of information that \emph{could} be transmitted from actions to future sensor states \cite{powerOrig, powerIntro}:
\begin{equation}\label{eq:power}
    \mathfrak{E}(e_t)= \max_{p(A_t)} I(A_t;S_{t+1}|e_t).
\end{equation}

Unfortunately, the channel capacity is notoriously expensive to compute and true empowerment often remains an ideal to be approximated in practice. There exist approximate solutions to this problem using neural networks \cite{nn2, nn1}, but this paper will present a novel approach, similar to that published in \cite{ildefons}, that offers the advantage of relative simplicity, while being similar in structure to a recently proposed implementation of intrinsic motivation \cite{Mila}.

\subsection{Variational Autoencoders}
Variational Autoencoders (VAE) are identity function approximators that map each input $x$ to a distribution $q(Z|x)$ over latent codes $z$, which is parameterized by $\phi$ and conditioned on the input data. To restore the original data, a single code $z$ is sampled from the distribution and subsequently decoded by another probability distribution $p(X|z)$ over possible inputs, which is parameterized by $\theta$ and conditioned on the sampled code \cite{VAEorig}.

The loss function, which is minimized during training, is the following \parencite{VAEorig}:
\begin{align}\label{eq:vae_loss}
    \mathcal{L}(x, \theta, \phi) = \mathbb{E}_{q_{\phi}(Z|x)}& [-\log{p_{\theta}(x|Z)}]\\ &+ KL[q_{\phi}(Z|x) \mid\mid p(Z)]. \notag
\end{align}

\section{Methods} \label{experiments}
\subsection{`Curious' Agent}\label{curAgent}
The information-gain-inspired agent that served both as a competitor and progenitor to the empowerment-inspired agent is itself based on an agent presented in \cite{Mila} by Klissarov et al. In the following, we will refer to this agent as the \emph{curious} agent. It is trained using the standard advantage actor-critic (A2C) algorithm (see Algorithm \ref{alg:a2c}), which we adapted from the `Torch\_AC' package \cite{torchAC}. Both the critic, as well as the actor are implemented using fully connected neural networks.

The main modification of the standard A2C algorithm was the introduction of a VAE, which is trained to compress the incoming sensor states. All VAEs featured in the following set of experiments were adapted from \cite{VAE}.

After every 128 frames, the VAE, as well as the actor and critic networks, are trained on the collected data (see Algorithm \ref{alg:a2c}). At every frame, the current sensor state is compressed and decoded by the VAE, yielding a loss in the form of \eqref{eq:vae_loss}. Similar to the approach in \cite{burda}, which uses random network distillation, a new sensor state that is completely unlike anything the agent has perceived before will be comparatively difficult to compress, since the VAE has not been trained on similar states. Therefore, the loss can be interpreted as signaling `novelty' and used as intrinsic reward:
\begin{equation} \label{eq:surprise}
    r^{in}_{t}(s_{t}, \theta, \phi) = \mathcal{L}(s_{t}, \theta, \phi).
\end{equation}

\noindent The resulting value for the intrinsic reward is first scaled by a factor, denoted $\beta$, and then added to the extrinsic reward to yield the total reward (see Algorithm \ref{alg:a2c}). Klissarov et al. used only the KL term of the loss as intrinsic reward, due to its similarity to Bayesian surprise \cite{Bayes, Mila}, while we obtained better results with the present approach. To ensure the fairness of the comparison with the empowerment-inspired agent below, we chose to work with the version that performed best. 
\begin{center}
\begin{algorithm}[t]
\SetAlgoLined
 \For{Episode $=0, 1, 2, \ldots$}{
  Initialize dataset $\mathcal{D}$ and insert $s_{0}$\;
  \For{$t=0, 1, 2, \ldots, T$}{
   take action $a_{t}$ and observe next state $s_{t+1}$ and extrinsic reward $r^{ex}_{t+1}$\;
   compute intrinsic reward  $r^{in}_{t} = \mathcal{L}(s_{t}, \theta, \phi)$\;
   store tuple $(s_{t+1}, a_{t}, r^{in}_{t+1}, r^{ex}_{t+1})$ in $\mathcal{D}$\;
   \If{$mod(t, N)$}{
   train the actor and critic on return $G_{t} = \sum_{t} r^{ex}_{t} + \beta r^{in}_{t}$\; 
   train the VAE on the collected states $s$ in $\mathcal{D}$\;
   initialize dataset $\mathcal{D}$ and insert $s_{t}$ in $\mathcal{D}$}
  }
 }
 \caption{The A2C learning algorithm with variational state encoding for intrinsic motivation. The only difference to standard A2C training is that, every $N$ frames ($N=128$, in our case), a VAE is trained on the previously collected sensory data. At \emph{every} frame, the VAE tries to compress the current sensor state $s_{t}$ and the resulting loss $\mathcal{L}(s_{t}, \theta, \phi)$ is used as intrinsic reward. The pseudocode is adapted from \cite{Mila}, and was the main source to inform the implementation of all agents presented in this paper.}\label{alg:a2c}
\end{algorithm}
\end{center}

\subsection{Empowerment Agent}\label{powerAgent}
The architecture of the curious agent above underwent a number of incremental changes that brought the intrinsic reward closer to signaling empowerment. The first step consisted in turning the VAE into a predictive model. Instead of using one and the same sensor state as both input and target, the state at time step $t+1$ served as a target vector, while the input was the sensor state at $t$. The VAE was thus turned into a \emph{deep variational information bottleneck} \cite{DVIB}, which we will henceforth refer to simply as a \emph{predictor}.

When future sensor states are used as targets, \eqref{eq:vae_loss} is turned into the following predictor loss function
\begin{align}\label{eq:pred_loss}
    \mathcal{L}_{\textrm{pred}}(x, x',\theta, \phi)= \mathbb{E}_{q_{\phi}(Z|x)}& [-\log p_{\theta}(x'|Z)]\\ &+ KL[q_{\phi}(Z|x) \mid\mid p(Z)], \notag
\end{align}

\noindent where $x$ refers to the input, $x'$ to the target, while the other variables are defined as in \eqref{eq:vae_loss}. The second modification was to add a second predictor, which differs from the first one in only one aspect. Instead of decoding the future sensor state $s_{t+1}$ from only the preceding state $s_{t}$, the second predictor takes in the preceding action $a_{t}$ as additional input. This second predictor will subsequently be referred to as the \emph{action-predictor}.

In a state of high empowerment, the action $a_{t}$ should bear more information about the subsequent sensor state $s_{t+1}$. To obtain an estimate of how much information about the future the pair $(s_{t}, a_{t})$ contains, over and above the information available from only $s_{t}$, one can compare the predictive performance of both predictors \cite{ildefons}. We chose the loss function as a performance metric, since it is already implicitly used as such during training. We then defined the intrinsic reward, which we want to be correlated with empowerment, as follows:
\begin{equation} \label{eq:quasi-power}
    r^{in}_{t} = \mathcal{L}_{\textrm{pred}}(s_{t}, s_{t+1},\theta_{2}, \phi_{2})-\mathcal{L}_{\textrm{pred}}((s_{t}, a_{t}), s_{t+1}, \theta_{1}, \phi_{1}).
\end{equation}

\noindent In \eqref{eq:quasi-power}, the two terms refer to the losses resulting from the prediction of $s_{t+1}$ by the action-predictor with parameters $\theta_{1}$, $\phi_{1}$ and predictor with parameters $\theta_{2}$, $\phi_{2}$. Both losses are calculated as defined in \eqref{eq:pred_loss}.

We now show that, under assumptions 1 through 3 below, the intrinsic reward of \eqref{eq:quasi-power} can be related to empowerment. As shown in \cite{DVIB}, the predictor loss of \eqref{eq:pred_loss} approximates a lower bound $L$ for the information bottleneck objective \cite[Eq.3]{DVIB} 
\begin{align}\label{eq:obj}
\mathcal{L}_{\textrm{pred}}(x,x',\theta,\phi) \approx L(x,x') \leq I(Z;X') - \alpha I(Z;X),   
\end{align}

\noindent where $x$ denotes the input, $x'$ the target, $Z$ the latent code, and $\alpha$ a Lagrange multiplier. Under assumptions 1 and 2 below, we get
\begin{align}\label{eq:deriv}
    r^{in}_{t} \approx &I(Z; S_{t+1}) - \alpha I(Z; S_{t}) \\
    &- I(Z; S_{t+1}) + \alpha I(Z; S_{t}, A_{t}). \notag
\end{align}

\noindent The right side of \eqref{eq:deriv} simplifies to 
\begin{align}\label{eq:info}
    \alpha I(Z;A_{t} \mid S_{t}). 
\end{align}

\noindent Then, by maximizing \eqref{eq:quasi-power}, the actor also maximizes \eqref{eq:info}. Under assumption 3 below, it thereby also maximizes \\$I(S_{t+1}; A_{t} \mid S_{t})$, which is a lower bound on the empowerment according to \eqref{eq:power}.

This argument rests on the following three assumptions:

\begin{enumerate}
    \item The latent variables of the two predictors can be treated as the same random variable $Z$.
    \item The difference of the losses in \eqref{eq:quasi-power} approximates the difference between the actual information bottleneck objective functions for both predictors according to \eqref{eq:obj}.
    \item Maximizing \eqref{eq:info} implies maximization of \\$I(S_{t+1}; A_{t} \mid S_{t})$.
\end{enumerate}

Assumptions 1 appears reasonable, since the latent variables for both predictors are of the same type and are used in the same way. Assumption 2 should be true at least when the bounds are similarly far away from the true values. Lastly, Assumption 3 should hold, if the encoder is forced to only inject information into $Z$ that is relevant for predicting $S_{t+1}$, as is the case if $Z$'s capacity is low enough.

Other than these changes, the training algorithm remained as it was for the curious agent (see Algorithm \ref{alg:a2c}).

There are multiple implementations of empowerment-like reward functions in the skill discovery literature \cite{gregor,eysenbach,sharma}. Replacing skills with mere actions in these publications could lead to an alternative to our approach. Note however, that the empowerment-like reward is not considered or evaluated as an exploration drive in these  publications. Note also that the empowerment estimation method in \cite{zhao} cannot be used for exploration, since the modelled channel form actions to sensors has a fixed noise term.

\section{Experiments} \label{realEx}
\subsection{Environments} \label{environments}
All environments used to experiment with the two agents above, were taken from the `MiniGrid' package \cite{GymMinigrid}. The environments are simple grid worlds, which contain a number of simple objects that can be manipulated by the agent. The agent only directly perceives a square area with a side length of seven tiles, which is located mostly in front of it.

\noindent All environments offer sparse rewards of magnitude one when a single `goal event' is reached, which always concludes the episode.

\subsubsection{MultiRoom-N3-S4-v0}
Here, the agent is placed in the first of three rooms of side length two. The three rooms are connected by doors, which are closed, but not locked. A goal tile is located in the third room and dispenses an extrinsic reward when reached by the agent. The relative locations of the three rooms is randomly reinitialized after each episode. We added this environment to the existing Minigrid-Multiroom configurations following the example of \cite{Mila}.

\subsubsection{DoorKey-8x8-v0}
This environment consists of two rooms separated by a wall, which taken together form a square of side length six. The agent is placed into the first room together with a key. It has to find the key, pick it up, carry it to a locked door in the wall, open the door and walk to a goal tile in the second room. It is only when it has reached the goal tile that it receives an extrinsic reward. The location of the wall and the door, the placement of agent and key in the first room, and the location of the goal tile in the second room vary randomly between episodes.

\subsubsection{KeyCorridor-S3R1-v0}
 The agent is placed in the middle of a corridor of length five and has to turn in order to open a door, behind which it finds a key. It needs to use the key to open a second door on the other side of the corridor. A ball is located behind this second door. Picking it up yields an extrinsic reward of one.

\subsection{Training}
The framework for the structure of the training process, including the handling of model parameter storage, command-line prompts, and log files, was based on the `RL-Starter-Files' package \cite{rl-starter-files}. The curious, the empowerment-inspired, as well as the A2C baseline agents were trained on a certain number of frames per run, depending on the environment. For every environment, the network parameters were first manually optimized before performing ten independent training runs (see Appendix A).

\section{Results}
The main metric of interest is the mean reward per episode, which is calculated using a sliding window over the last eight episodes. Each figure below plots the mean reward per episode, averaged over the ten independent runs, along with its standard deviation (shaded area).

For each agent and environment, the frame number at which the mean reward per episode first reached a value of one, was recorded, yielding ten values per agent. These values will subsequently be referred to as the \emph{time-to-success} of an agent. If an agent never solves a task, the time-to-success is set to ten million frames. By analyzing the time-to-success data, the question of whether one agent learns more quickly than another can be statistically addressed using a one-sided $t$-test. A difference is considered significant if $p<0.05$, where $p$ denotes the $p$-value. The direction of the hypotheses was informed by the plots of the training progress. Descriptive statistics for all agents and environments are summarized in Table \ref{tab:allstats}.

\subsection{MultiRoom-N3-S4-v0}\label{N3S4Results}
\begin{figure}[t]
\centering
\includegraphics[width=.4\textwidth]{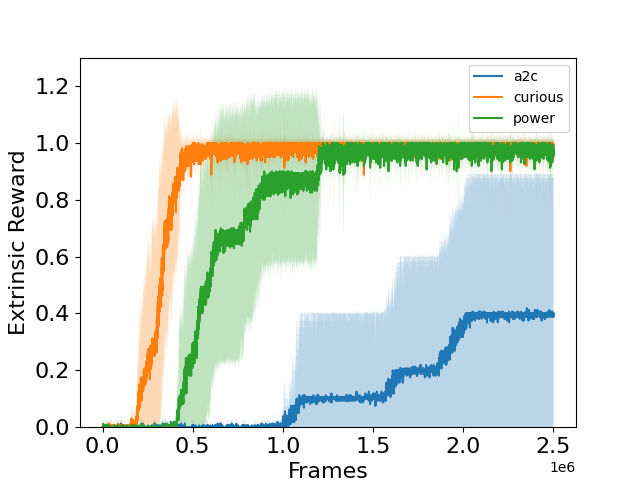} 
\caption[Plot of Results for MultiRoom-N3-S4-v0]{Training results for the MultiRoom-N3-S4-v0 environment. The curious agent (orange) performs best, followed by the empowerment-inspired agent (green), here labelled `power', and lastly the A2C baseline (blue).}
\label{fig:N3S4}
\end{figure}

The time-to-success of the curious agent was significantly shorter compared to that of the A2C baseline and the empowerment-inspired agent. The empowerment-inspired agent also solved the task more quickly than the baseline (see Table \ref{tab:ttestN3S4} and Fig. \ref{fig:N3S4}).
\begin{center}
 \begin{table}[b]
 \centering
 \begin{tabular}{||c|c c c||} 
 \hline
     & A2C & curious & power\\ [0.5ex] 
 \hline\hline
 curious & \makecell[l]{$t=4.65$, \\ $p<0.001$} & --- & \makecell[l]{$t=4.19$,\\ $p<0.001$} \\ 
 \hline
 power & \makecell[l]{$t=4.39$, \\ $p<0.001$} & --- & --- \\ 
 \hline
\end{tabular}
\caption{This table includes the $t$ and $p$-values of the directed $t$-tests conducted on the time-to-success data from the MultiRoom-N3-S4-v0 environment. For any cell in the table, the alternative hypothesis is that the `row-agent' learns significantly faster than `column-agent'. `Power' refers to the empowerment-inspired agent.} \label{tab:ttestN3S4}
\end{table}
\end{center}

\subsection{DoorKey-8x8-v0}
\begin{figure}[b]
\centering
\includegraphics[width=.4\textwidth]{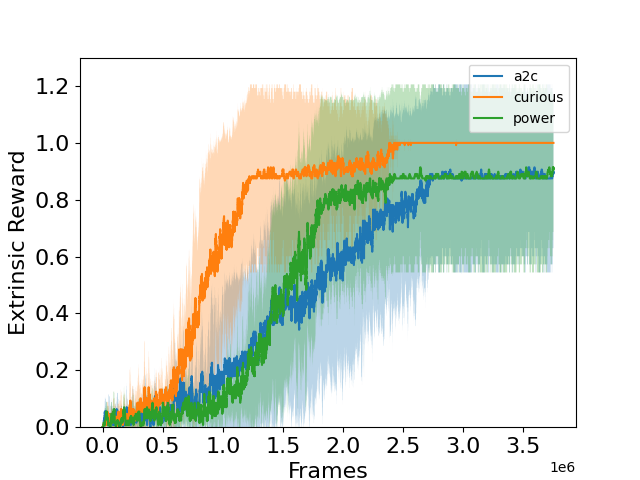} 
\caption[Plot of Results for DoorKey-8x8-v0]{Training results for the DoorKey-8x8-v0 environment. The curious agent (orange) seems to perform best, followed by the empowerment-inspired agent (green), here labelled `power', and lastly the A2C baseline (blue). These intuitions are in part statistically confirmed, however only after outlier values from the third run are removed.}
 \label{fig:DoorKey}
\end{figure}

In terms of pure time-to-success there was no significant difference between the three agents (see Table \ref{tab:ttestDoorKey} and Fig. \ref{fig:DoorKey}).

Interestingly, closer inspection of the data reveals that for both the baseline and the empowerment agent, the third run was the only one during which they failed to solve the task, thus resulting in one outlier time-to-success value of $1e7$ for the two agents. Even for the curious agent, the third run was the only trial during which it solved the task only after more or less $2.5e6$ frames, instead of after around $1.5e6$ frames. After removing the data from the third run and repeating the statistical analysis, we found that the curious agent learned more quickly than both the A2C baseline and the empowerment-inspired agent, while there was no difference in terms of time-to-success between the empowerment-inspired agent and the A2C baseline (see Table \ref{tab:ttestDoorKey}).

\begin{center}
 \begin{table}[t]
 \centering
 \begin{tabular}{||c|c c c||} 
 \hline
     & A2C & curious & power \\ [0.5ex] 
 \hline\hline
 curious & \makecell[l]{$t=1.68$, \\$p=0.067$, \\$t*=2.29$, \\$p*=0.021$} & --- & \makecell[l]{$t=1.52$, \\$p=0.085$, \\$t*=2.26$, \\ $p*=0.024$}\\ 
 \hline
 power & \makecell[l]{$t=0.11$, \\$p=0.46$, \\$t*=0.64$, \\$p*=0.27$} & --- & --- \\ 
 \hline
\end{tabular}
\caption{This table includes the $t$ and $p$-values of the directed $t$-tests conducted on the time-to-success data from the DoorKey-8x8-v0 environment. For any cell in the table, the alternative hypothesis is that the `row-agent' learns significantly faster than the `column-agent'. After the outlier time-to-success values from the third run were removed, the tests were repeated (results marked with a $*$).} \label{tab:ttestDoorKey}
\end{table}
\end{center}

\subsection{KeyCorridor-S3R1-v0}
\begin{figure}[b]
\centering
 \includegraphics[width=.4\textwidth]{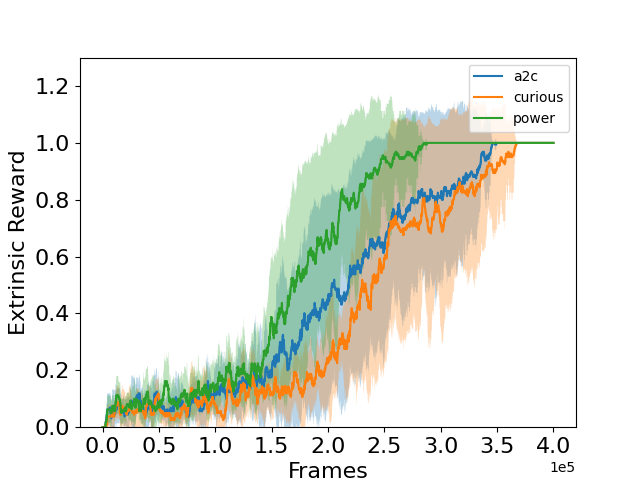} 
\caption[Plot of Results for KeyCorridor-S3R1-v0]{Training results for the KeyCorridor-S3R1-v0 environment. As suggested by the plots and confirmed by statistical analyses of the time-to-successs, the A2C baseline (blue) and curious agent (orange) perform equally, while the empowerment-inspired agent (green), labelled `power', learns more quickly than both.}
  \label{fig:KeyCorridorResults}
\end{figure}

The empowerment-inspired agent learned more quickly than both the A2C baseline and the curious agent. However, there was no significant difference in terms of time-to-success between the curious agent and the A2C baseline (see Table \ref{tab:ttestKeyCorridor} and Fig. \ref{fig:KeyCorridorResults}).

\begin{center}
 \begin{table}[h!]
 \centering
 \begin{tabular}{||c|c c c||} 
 \hline
     & A2C & curious & power \\ [0.5ex] 
 \hline\hline
 A2C & --- & \makecell[l]{$t= 1.1$, \\$p=0.28$} & --- \\ 
 \hline
 power & \makecell[l]{$t=1.86$, \\$p=0.041$} & \makecell[l]{$t=3.5$, \\$p=0.001$} & --- \\ 
 \hline
\end{tabular}
\caption{This table includes the $t$ and $p$-values of the directed $t$-tests conducted on the time-to-success data from the KeyCorridor-S3R1-v0 environment. For any cell in the table, the alternative hypothesis is that the `row-agent' learns significantly faster than the `column-agent'.} \label{tab:ttestKeyCorridor}
\end{table}
\end{center}

\begin{center}
 \begin{table}[h!]
 \centering
 \begin{tabular}{||l c c c c||} 
 \hline
  & & MultiRoom & DoorKey &  KeyCorridor \\ [0.5ex] 
 \hline\hline
 A2C & \makecell[l]{M:\\SD:} & \makecell{6.7e6 \\4.3e6} & \makecell{2.8e6 \\2.6e6} & \makecell{2.5e5 \\0.7e5} \\
 \hline
 curious & \makecell{M:\\SD:} & \makecell{3.3e5 \\0.8e5} & \makecell{1.2e6 \\0.5e6} & \makecell{2.8e5 \\0.6e5} \\ 
 \hline
 power & \makecell[l]{M:\\SD:} & \makecell{6.7e5 \\2.4e5} & \makecell{2.7e6 \\2.9e6} & \makecell{2.1e5 \\0.4e5} \\ 
 \hline
\end{tabular}
\caption{Mean (M) and standard deviation (SD) of the time-to-success for all agents and environments.} \label{tab:allstats}
\end{table}
\end{center}

\section{Discussion} \label{discussion}
This work, albeit with a slightly modified approach, partly replicates the finding by Kilssarov et al. \cite{Mila} that variational state encoding can be leveraged to build an intrinsically motivated agent, which outcompetes classical A2C. On the other hand, the benefit provided by the intrinsic reward depends on the task at hand and, in some cases, the surprise signal does not lead to significant improvements. Further, the alternative intrinsic reward function, inspired by the notion of empowerment, is similarly successful at counteracting sparsity of rewards and might therefore present a possible, scalable, empowerment approximation, which merits further investigation. This might be considered a hint that empowerment can drive exploration and that it may do so to a similar extent as an analogous intrinsic motivation inspired by the notion of information gain. However, significant assumptions and approximations are made in both agents and more research beyond these preliminary results is needed to establish whether empowerment maximization consistently leads to exploration.

\appendices
\section{Network Architectures and Training Parameters}

\begin{center}
 \begin{table}[b]
 \centering
 \begin{tabular}{||c c c c||} 
 \hline
 Agent & Hidden units & Latent dimensions & $\beta$ \\ [0.5ex] 
 \hline\hline
 curious & 512, 512, 265 & 256, 256, 128 & 2e-4, 1e-4, 1e-4 \\ 
 \hline
 power & 512, 32, 265 & 256, 16, 128 & 1e-4, 1e-4, 0.125e-4 \\
 \hline
\end{tabular}
\caption{Network parameters of the VAE/predictor models and $\beta$ values for all environments. In every cell, the first value corresponds to the MultiRoom-N3-S4-v0 environment, followed by the values for the DoorKey-8x8-v0  and KeyCorridor-S3R1-v0 environments, respectively.}
\end{table}\label{tab:params}
\end{center}

The observation arrays provided by the environment are converted into binary form, omitting the color information, before being forward propagated through the VAE/predictor. A given binary representation features a $1$ at index $[i, j, k, l]$, if and only if the original observation features object $k$ in state $l$, where $k$ and $l$ are integers, at position $(i, j)$ and a $0$ otherwise. The encoder and decoder of all VAEs/predictors feature one hidden layer and encoder and decoder of the same VAE/predictor always have the same number of hidden units (see Table V). All layers are densely connected. We used the ReLU activation function for the hidden layers, a linear activation function for the bottleneck layer, and a sigmoid for the output of the decoders. All models were trained with a learning rate of $0.001$ and the \emph{Adam} optimizer.

\section*{Acknowledgements}
We are much indebted to the entire research team of Araya Inc. for their constructive criticism, in particular Dr. Yen Yu.
\printbibliography
\end{document}